\documentclass[times, review, 10pt]{elsarticle}
\usepackage{amsmath}
\usepackage{bm}
\usepackage{graphicx}
\usepackage{tabularx}
\usepackage{booktabs}
\usepackage{geometry}
\usepackage{amssymb}
\usepackage{url}
\usepackage[hidelinks]{hyperref}
\usepackage{cleveref}
\usepackage{multirow}
\usepackage{multicol}
\usepackage{pifont}
\usepackage{colortbl}
\usepackage[table, dvipsnames]{xcolor}
\usepackage{mathtools}
\usepackage{upgreek}
\usepackage{array}
\usepackage{makecell}
\usepackage{algorithm}
\usepackage{algpseudocode}

\usepackage{amssymb}
\usepackage{amsmath}

\journal{Pattern Recognition}

\begin{document}

\begin{frontmatter}

\title{Hierarchical Mutual Distillation for Multi-View Fusion: Learning from All Possible View Combinations} 

\author[inst1]{Jiwoong Yang}
\ead{jiwoong0412@hanyang.ac.kr}
\author[inst1,inst2,inst3]{Haejun Chung\corref{cor1}}
\ead{haejun@hanyang.ac.kr}
\author[inst4,inst5,inst6]{Ikbeom Jang\corref{cor1}}
\ead{ijang@hufs.ac.kr}
\cortext[cor1]{Corresponding authors}

\affiliation[inst1]{organization={Department of Artificial Intelligence Semiconductor Engineering, Hanyang University},
            addressline={222, Wangsimni-ro, Seongdong-gu},
            city={Seoul},
            postcode={04763},
            country={Republic of Korea}}
\affiliation[inst2]{organization={Department of Artificial Intelligence, Hanyang University},
            addressline={222, Wangsimni-ro, Seongdong-gu},
            city={Seoul},
            postcode={04763},
            country={Republic of Korea}}
\affiliation[inst3]{organization={Department of Electronic Engineering, Hanyang University},
            addressline={222, Wangsimni-ro, Seongdong-gu},
            city={Seoul},
            postcode={04763},
            country={Republic of Korea}}
\affiliation[inst4]{organization={Division of Computer Engineering, Hankuk University of Foreign Studies},
            addressline={81, Oedae-ro, Mohyeon-eup, Cheoin-gu},
            city={Yongin},
            postcode={17035},
            country={Republic of Korea}}
\affiliation[inst5]{organization={Division of AI Data Convergence, Hankuk University of Foreign Studies},
            addressline={81, Oedae-ro, Mohyeon-eup, Cheoin-gu},
            city={Yongin},
            postcode={17035},
            country={Republic of Korea}}
\affiliation[inst6]{organization={Division of Language and AI, Hankuk University of Foreign Studies},
            addressline={107, Imun-ro, Dongdaemun-gu},
            city={Seoul},
            postcode={02450},
            country={Republic of Korea}}

\begin{abstract}
Multi-view learning often struggles to effectively leverage images captured from diverse angles and locations. Learning methods for unstructured multi-view images remain largely underexplored. We propose a novel Hierarchical Mutual Distillation for Multi-View Fusion (HMDMV) method, which can handle both structured and unstructured multi-view scenarios. 
It makes predictions utilizing all possible view combinations: single view, partial multi-view, and full multi-view. The method generates predictions for each view combination and then applies hierarchical mutual distillation to enhance inter-view consistency. An uncertainty-based weighting mechanism further refines the fusion process by adjusting the influence of each view combination according to its prediction confidence, reducing the impact of low-confidence views. 
Extensive experiments on large-scale structured and unstructured datasets demonstrate that HMDMV consistently achieves state-of-the-art classification accuracy. 
Another unique advantage of HMDMV is that it provides improved flexibility in inference, allowing for more or fewer view counts in inference than those used in training without additional processing.
We also provide a light version with reduced training cost by designing an efficient strategy that randomly samples subsets of view combinations during each training iteration.
These results highlight HMDMV’s robustness in real-world settings where view availability is variable or incomplete. The code is available at \url{https://github.com/labhai/HMDMV}.



\end{abstract}

\begin{keyword}
Multi-view learning \sep Hierarchical mutual distillation \sep Uncertainty-aware fusion \sep Flexible multi-view inference \sep Image classification
\end{keyword}

\end{frontmatter}


\section{Introduction}
\label{sec:intro}
\begin{figure}[t]
\centering
\includegraphics[width=0.6\columnwidth]{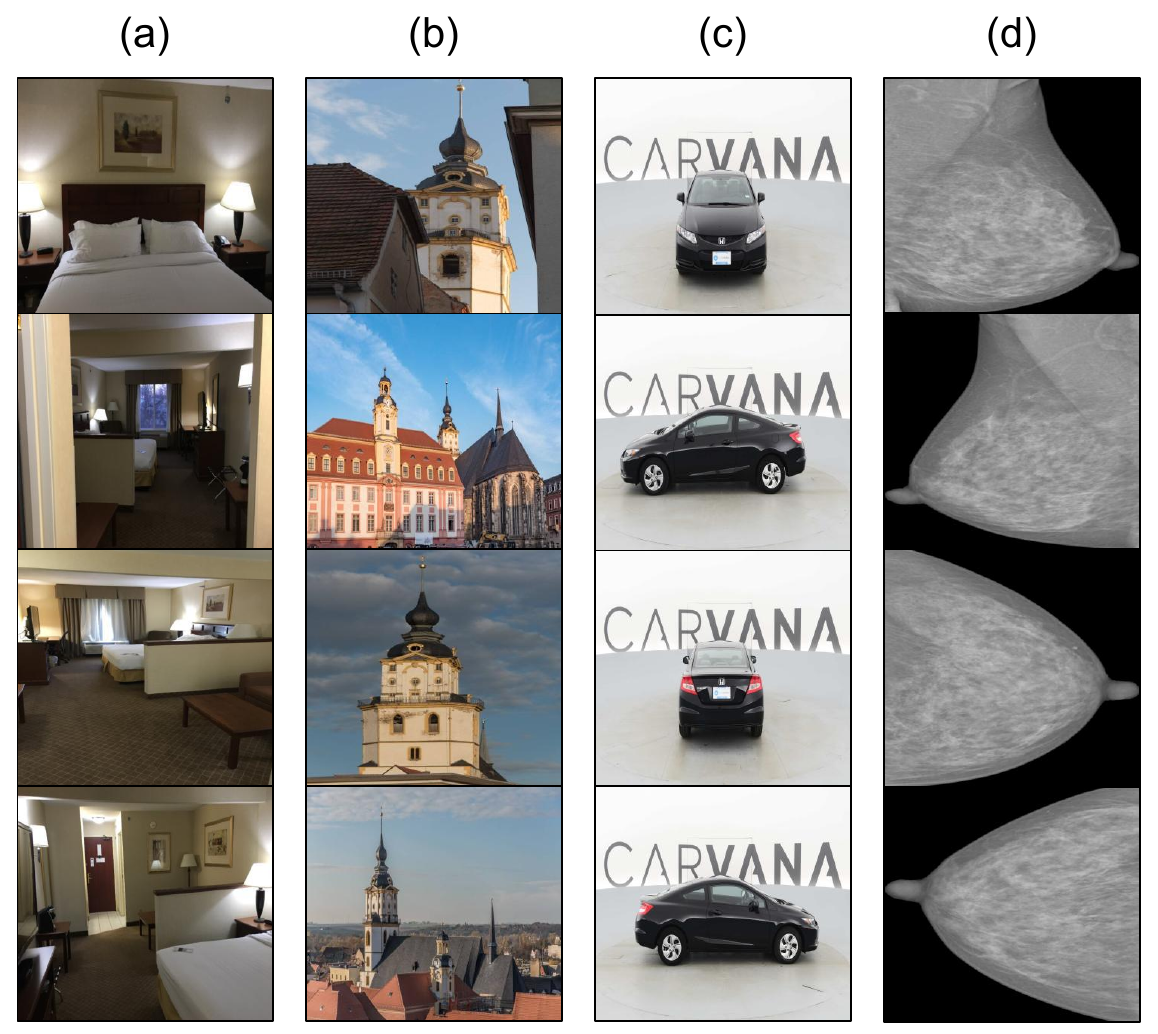}
\caption{Examples of multi-view datasets. (a) Hotels-8k~\cite{kamath20212021} consists of hotel room interior images captured from various angles, distances, and lighting conditions. (b) Google Landmarks Dataset v2~\cite{weyand2020google} includes landmark images captured from a range of perspectives, such as ground-level, aerial, and oblique views. (c) Carvana~\cite{carvana-image-masking-challenge} contains vehicle images consistently captured from predefined viewpoints (e.g., front, side, and rear) in a controlled indoor environment. (d) VinDr-Mammo~\cite{nguyen2023vindr} provides full-field digital mammography images, consisting of standard screening views (CC and MLO) for abnormality detection.}
\label{figure_1}
\end{figure}

Multi-view learning integrates complementary information captured from multiple viewpoints of an object or a scene, enabling more accurate and robust visual understanding. Image data is generally represented as 2D projections, capturing limited planar information from specific viewpoints. This representation can lead to information loss regarding the 3D structure and context of an object. Leveraging multiple views mitigates this limitation by capturing diverse and complementary visual cues across viewpoints. Thus, multi-view fusion applies to various computer vision fields. In medical imaging analysis, for instance, structured multi-view data from predefined angles is employed to achieve precise diagnoses. Specifically, datasets for chest diagnosis typically consist of Frontal and Lateral X-ray views~\cite{zhu2021mvc, kim2023chexfusion}, while mammography for breast cancer screening utilizes Craniocaudal (CC) and Mediolateral Oblique (MLO) views~\cite{xia2023neural, manigrasso2025mammography}. In contrast, in non-medical fields, some less structured cross-view tasks captured from diverse angles are often used, such as for 3D object recognition~\cite{su2015multi, feng2018gvcnn} and action recognition~\cite{xiao2019action}. Recent studies have also extended the single-view problem to an unstructured dataset composed of various angles and environments that are not fixed~\cite{black2024multi}. \Cref{figure_1} shows representative examples of multi-view image datasets.

\begin{figure}[t]
\centering
\includegraphics[width=0.7\columnwidth]{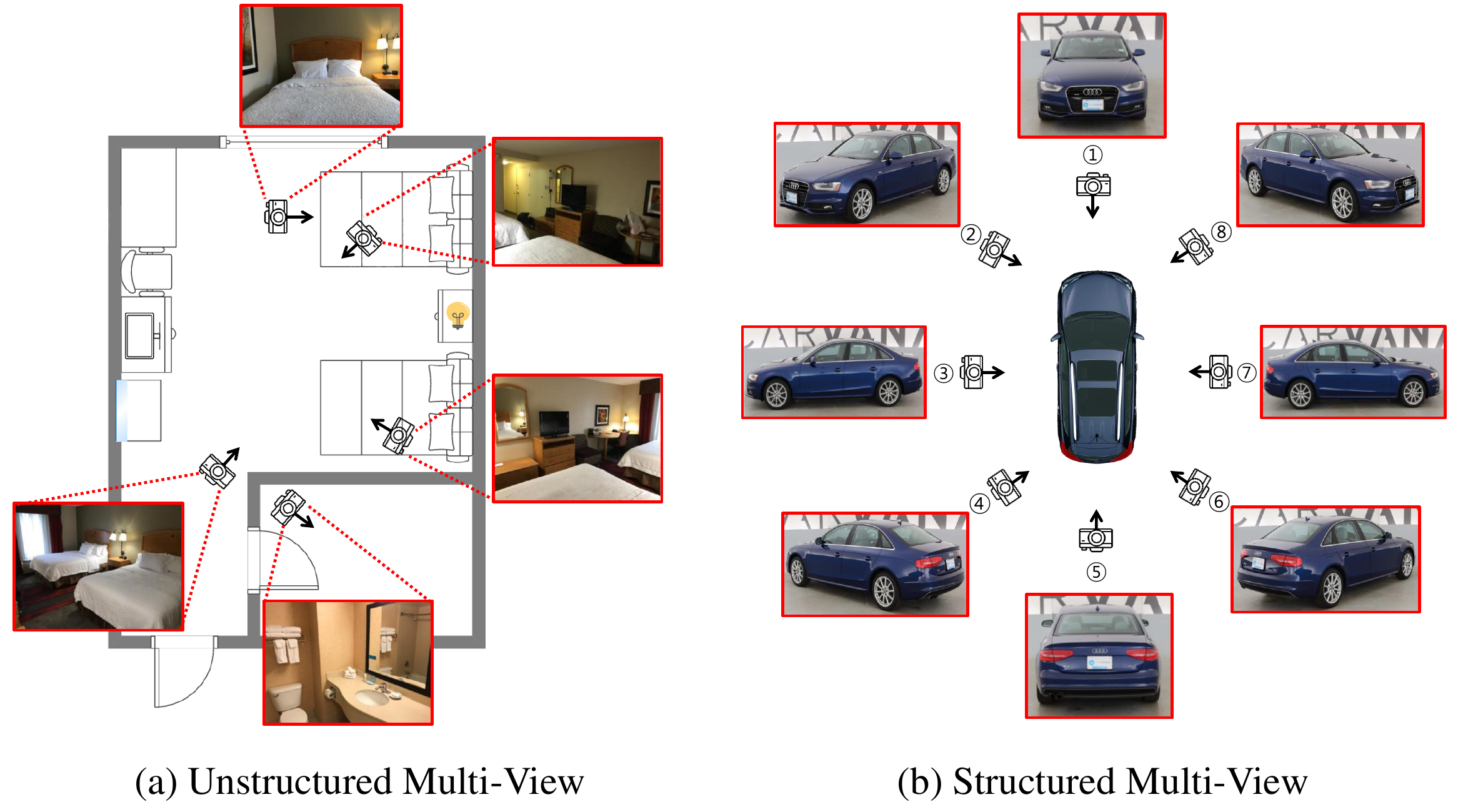}
\caption{Comparison of structured and unstructured multi-view setups. (a) Unstructured multi-view setup is taken from various locations, angles, and lighting conditions without predefined camera positions, providing diverse perspectives of the sample. (b) Structured multi-view setup is consistently captured from a fixed set of predefined camera positions (e.g., front, side, rear, and oblique angles) around the sample, ensuring consistent viewpoints and the same number of views.}
\label{figure_2}
\end{figure}

Previous multi-view fusion methods have focused on combining features extracted from each view to generate a final prediction. Early approaches, for instance, averaged features or aggregated independent predictions from each view. While these methods demonstrated improvements over single-view baselines, they often failed to capture cross-view inconsistencies and interactions fully, reducing the consistency of the predictions. Recently, approaches using a hybrid CNN-Transformer network~\cite{dosovitskiy2020image} have been proposed to perform feature fusion across views or to apply mutual distillation~\cite{black2024multi, guo2024multi} between single-view and multi-view predictions to strengthen the relationships among views. While mutual distillation strategies can improve cross-view consistency by facilitating bidirectional knowledge transfer between single-view and fused multi-view predictions, most existing designs primarily rely on a binary distillation mechanism. This implicitly assumes that directly aligning single-view outputs with the fully fused representation is sufficient. However, in unstructured settings, the fused representation may aggregate heterogeneous observations, including noisy or occluded views, which can introduce ambiguity during mutual learning. Moreover, by bypassing intermediate subset interactions, prior methods underutilize complementary information in partial view sets, limiting robustness across varying view availability. Consequently, these methods often provide limited inference flexibility when view availability is incomplete or varies across samples. Furthermore, these do not sufficiently account for the uncertainties of each view, thus limiting their ability to ensure prediction consistency fully. In contrast to structured multi-view fusion, as illustrated in \Cref{figure_2}, these challenges are amplified in unstructured multi-view fusion due to the arbitrary number of images and non-fixed viewpoints.

In this work, we propose the Hierarchical Mutual Distillation for Multi-View Fusion (HMDMV) method. We demonstrate that it outperforms existing approaches in both unstructured and structured multi-view settings. It explicitly considers the cross-view relationships among all views and enhances prediction performance.
We define three levels of view combinations, which will be used throughout the paper when explaining the combination or fusion: \textit{Single-view} (individual views), \textit{Partial Multi-view} (combinations of a subset of views), and \textit{Full Multi-view} (combination of all views). 
First, we extend a hybrid CNN–Transformer architecture to generate predictions from all possible view combinations, enabling the model to leverage complementary information across different subsets.
We then fuse these predictions using uncertainty-based weighting and perform hierarchical mutual knowledge distillation from single-view and partial multi-view predictions to the full multi-view prediction. To further address the computational overhead of considering all possible view combinations, we introduce an efficient repeated random subset sampling strategy that substantially lowers training cost while preserving performance. Our contributions are summarized below:
\begin{itemize}
    \item We propose a hierarchical mutual distillation approach that leverages all possible view combinations for multi-view fusion, enabling more stable and consistent predictions.

    \item Our method improves the prediction performance of view combinations by dynamically weighting them based on their prediction confidence through an uncertainty-aware strategy.
    
    \item The method supports flexible inference across varying view counts and configurations, enabling practical deployment in real-world multi-view scenarios.
    
    \item It achieves competitive or state-of-the-art performance on both structured and unstructured datasets.
    
\end{itemize}

\section{Related works}
\label{sec:related_work}
Multi-view learning has recently evolved to encompass diverse methodologies across various tasks, such as multi-view feature selection, multi-view clustering, and multi-view classification. Multi-view feature selection seeks a compact set of discriminative features from high-dimensional multi-view data, reducing redundancy while preserving view-specific and shared structures. Several approaches jointly optimize graph structure learning with feature selection through adaptive structure learning and iterative inference \cite{zhang2024efficient}, consensus pseudo-labels derived from multiple views \cite{cao2024structure}, and graph discrepancy learning that captures both global and local structures in kernel space \cite{xu2025multi, xu2025multi2}. Multi-view clustering aims to partition data into groups by leveraging both consensus and complementary information across views without label supervision. Recent studies have explored collaborative similarity fusion with consistency recovery for incomplete scenarios \cite{jiang2025collaborative} and sample-level adaptive fusion with contrastive learning \cite{yang2025deep}. Multi-view classification focuses on learning that fuses information from multiple views to improve prediction accuracy. Some studies have investigated adaptive view weighting through regression-based frameworks \cite{jiang2021robust, jiang2022robust}, anchor-based collaborative fusion that automatically adjusts view importance \cite{jiang2023adaptive}, and deep learning approaches combining feature extractors with large-margin classifiers or contrastive learning \cite{xie2023deep, li2025multi}.
In this work, we focus on multi-view image classification and introduce a novel multi-view fusion strategy. In the following, we review related work on multi-view fusion methods, uncertainty-based weighting, and mutual distillation learning, which are relevant to our proposed method.

\subsection{Multi-view fusion methods}
In multi-view learning, feature fusion methods are often categorized by the stage at which information from each view is combined: early fusion, late fusion, and score fusion.

\textbf{Early fusion} merges low-level features from each view before model training, letting the model treat them as a single-view input. Some approaches perform learning on fused representations by integrating shallow feature maps extracted from each view of a multi-view image~\cite{zhu2021mvc, sun2019multi}. Early fusion is also used in multimodal tasks, for example by integrating audio–visual inputs into a unified representation~\cite{zhang2025audio}. However, as this fusion occurs before feature extraction, it can introduce irrelevant information and dilute task-relevant features~\cite{zhang2021deep}.
\textbf{Late fusion} is widely used in multi-view and multimodal data, learning features independently for each input before combining them. Early studies applied simple concatenation~\cite{seeland2021multi} or pooling~\cite{su2015multi}. As research advanced in 3D object recognition and multi-modal learning, more advanced fusion methods emerged~\cite{wei2022learning, chen2021mvt, tang2024itfuse}. Recent multi-view fusion strategies also have expanded to integrate deep feature extractors with discriminative classifiers~\cite{xie2023deep, li2025multi}.
\textbf{Score fusion} combines probability distributions from each view to produce a final prediction. Among the proposed approaches, some methods pool single-view class distributions element-wise~\cite{seeland2021multi}, while others adopt the Dirichlet distribution~\cite{han2022trusted} to incorporate uncertainty into the process of combining predictions from multiple views. A hybrid fusion approach, introduced by Black et al.~\cite{black2024multi}, combines late fusion and score fusion by leveraging both the predictions obtained from fused features across views and the fusion of individual view predictions. Motivated by this approach, we propose extending it to perform fusion over all possible view combinations, enabling the model to capture complementary information from partial multi-view subsets.
\subsection{Uncertainty-based weighting}
Uncertainty measures the confidence of a deep learning model's predictions~\cite{gal2016dropout}. There has been sustained interest in leveraging uncertainty information during training to enhance model robustness and stability.
Uncertainty-based weighting dynamically adjusts learning weights based on predictive uncertainty. For instance, the double-uncertainty weighted method~\cite{wang2020double} simultaneously considers segmentation and feature uncertainty to weight the consistency loss, thereby reducing the impact of unreliable predictions and improving semi-supervised segmentation performance. In multi-task learning, Soft Optimal Uncertainty Weighting (UW-SO)~\cite{kirchdorfer2024analytical} derives task weightings analytically by interpreting gradient variance as task-specific uncertainty, yielding more stable and generalizable optimization. Homoscedastic uncertainty-based weighting~\cite{kendall2018multi} instead models task-dependent observation noise and incorporates it into the loss, enabling automatic balancing of heterogeneous tasks such as regression and classification.
However, in multi-view learning tasks, leveraging the uncertainty of each view often increases computational complexity, and obtaining precise measurements remains challenging. These limitations highlight the necessity of developing stable and efficient uncertainty modeling strategies.
\subsection{Mutual distillation learning}
Knowledge Distillation~\cite{hinton2015distilling} is a technique that facilitates the transfer of knowledge from a large, more complex model (teacher) to a smaller model (student). Conventionally, the student model's loss includes an extra divergence term that quantifies the discrepancy between the predictions of the teacher and the student model. This approach remains an active area of research, with recent studies further advancing knowledge distillation for classification task~\cite{zhu2024dynamickd, sun2025explainability}.
Mutual Distillation enables bidirectional knowledge transfer, transforming the one-way process of knowledge distillation into a mutually enhancing approach. Deep Mutual Learning (DML)~\cite{zhang2018deep} trains multiple models simultaneously, enabling each network to serve as a teacher for the others. Feature Fusion Learning (FFL)~\cite{kim2021feature} employs an online distillation strategy where sub-network and fused classifiers are jointly trained. Dense Cross-layer Mutual-distillation (DCM)~\cite{yao2020knowledge} leverages dense layer-to-layer interactions to enable mutual distillation.
Recently, distillation methods for multi-view learning have been studied. ViewsKD (VKD)~\cite{porrello2020robust} facilitates the transfer of diverse visual knowledge across multiple views of an object. MVC-Net~\cite{zhu2021mvc} applies self-distilling mimicry loss to align paired view predictions, enhancing class probability consistency. MV-HFMD~\cite{black2024multi} adopts mutual distillation to align the predictions of single views and fused multi-view. While this enables bidirectional knowledge transfer, this design enforces direct alignment between single-view and fused multi-view representations, bypassing intermediate contexts. In unstructured settings where the fused representation may be compromised by outliers, such binary alignment may propagate noise bidirectionally. To address this gap, we propose performing mutual distillation across all view combinations, using partial predictions as intermediate representations.
\section{Method}
\label{sec:method}
\begin{figure}[t]
\centering
\includegraphics[width=1.0\columnwidth]{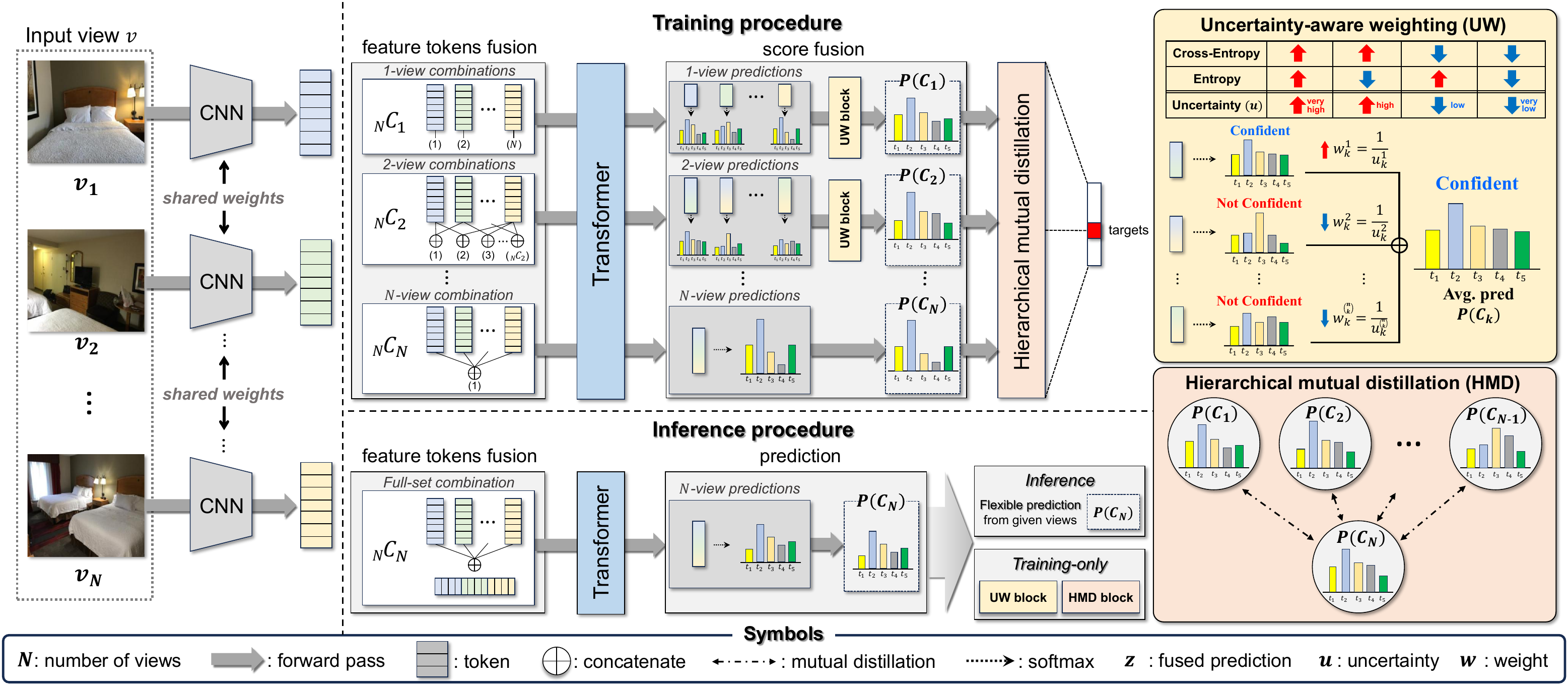}
\caption{Overview of HMDMV framework. Given $N$ input views, feature tokens are extracted through a hybrid CNN-Transformer network with shared weights. During training, all possible $k$-view combinations are generated and processed through the Uncertainty-aware Weighting (UW) block and Hierarchical Mutual Distillation (HMD) block.  The UW block assigns higher weights to confident predictions based on cross-entropy and entropy. The HMD block performs bidirectional knowledge distillation by aligning the predictions of each combination set $P(C_k)$ for $1\leq k < N$ with the full multi-view prediction $P(\mathcal{C}_N)$. During inference, only the full-set combination is used to produce the final prediction.}
\label{figure_3}
\end{figure}
We propose a method that, unlike conventional approaches that rely on the independent learning of single views or simple concatenation, leverages all possible view combinations to generate comprehensive multi-view representations. For each set of generated combinations, we perform an uncertainty-weighted score fusion and then conduct a hierarchical mutual distillation, ultimately refining the full multi-view predictions. \Cref{figure_3} provides an overview of our proposed method.

\subsection{All-possible view combinations fusion}
\label{sec:All_comb}
We extend a hybrid CNN-Transformer network~\cite{dosovitskiy2020image} to generate feature tokens for each view and fuse all possible combinations. Hybrid CNN-Transformer networks have demonstrated their effectiveness in computer vision tasks. CNNs excel at extracting local spatial features, while Transformers utilize self-attention mechanisms to capture global dependencies. In multi-view scenarios, these advantages prove effective for learning information from multiple perspectives and fusing features.

In this network, the shared-weight CNN transforms each input view \( v \) into a feature map \(f_v \in \mathbb{R}^{h \times w \times c}\), where \((h, w)\) denotes the downsampled spatial resolution and \(c\) is the number of channels. The \(f_v\) is flattened along the spatial dimension to produce \(S=hw\) vectors $\{f_{v,l}\}_{l=1}^S$, each of size \(c\). A linear projection matrix \(W \in \mathbb{R}^{c \times d}\) is applied to these vectors, generating a sequence of \(S\) tokens. A learnable positional encoding \( \mathbf{E}_{\text{pos}} \in \mathbb{R}^{S \times d}\) is added to incorporate spatial position information:
\begin{equation}
    \mathcal{E}(v) = [f_{v,1}W; f_{v,2}W; \dots; f_{v,S}W] + \mathbf{E}_{\text{pos}}
  \label{eq:feature_token}
\end{equation}
where \(\mathcal{E}(v)\) denotes the token sequence of view $v$, and \(f_{v,l}\) refers to the \(l\)-th spatial feature.
Given \( n \) input views, we define \(\mathcal{C}_k\) as the set of all possible \(k\)-view subsets:
\begin{equation}
    \mathcal{C}_k = \bigl\{\,C_k^i \; | \; C_k^i \subset \{v_1,\dots,v_n\},\; |C_k^i|=k\}
    \label{eq:view_subsets}
\end{equation}
where $i\!\in\!\{1, 2, \dots , \binom{n}{k}\}$.
Here, \( k\!=\!1 \) represents each single view; $1\!<\!k\!<\!n$ denotes partial multi-view combinations; and \( k\!=\!n \) represents the full multi-view. 
Each element \(C_k^i\!\in\!\mathcal{C}_k\) is a specific subset of \(k\) distinct views,
\(\displaystyle C_k^i\!=\!\{\,v_{i_1},\,v_{i_2},\,\dots,\,v_{i_k}\}\).
For each subset \(C_k^i\), we concatenate the feature tokens of its \(k\) views:
\begin{equation}
    \mathcal{E}(C_k^i) =
    \Bigl[\mathcal{E}(v_{i_1}),\; \mathcal{E}(v_{i_2}),\; \dots,\; \mathcal{E}(v_{i_k})\Bigr]
    \label{eq:combined_feature_token}
\end{equation}
Consequently, for combination sets $\mathcal{C}_k$, we obtain \(\binom{n}{k}\) such concatenated token sequences. The \(\mathcal{E}(C_k^i)\) are then passed through the transformer:
\begin{equation}
    p_k^i = T\bigl(\mathcal{E}(C_k^i)\bigr)
    \label{eq:prediction}
\end{equation}
where \( T \) represents the transformer block, and \(p_k^i\) denotes the logits or prediction distribution for the subset \(C_k^i\). In this way, we collect the predictions for \emph{all} combinations across \(k = 1, 2, \dots, n\). Therefore, this approach considers all possible ways that the features of the views can be fused, enabling the model to learn the dependencies and contextual relationships between views comprehensively. However, using all possible combinations without constraints leads to a significant increase in computational complexity as the number of views grows. To address this, we introduce a partial sampling-based strategy in \Cref{sec:scalable_training}.
\subsection{Uncertainty-weighted strategy}
\label{sec:UW}
To perform mutual distillation among different view combination sets, we propose an approach that assigns weights based on each element’s prediction uncertainty. The uncertainty of predictions from various view combinations is crucial.
While full multi-view outputs a single prediction, single-view and partial multi-view each generate multiple predictions, one per element in the combination set.

By integrating both entropy and cross-entropy, our approach captures high-confidence errors and thus assigns a more accurate overall uncertainty during training.
Entropy quantifies the distributional uncertainty of the prediction~\cite{shannon1948mathematical, distelzweig2024entropy}. However, when uncertainty is computed solely using entropy, a potential issue arises where the prediction may exhibit high confidence in an incorrect class while still reflecting a low uncertainty. Cross-entropy complements this by measuring how closely the predicted distribution aligns with the ground truth label, enabling the model to penalize incorrect confident predictions.

We define the uncertainty $u(p, y)$ of a prediction distribution $p$ for the ground truth label $y$ as the sum of the Shannon entropy $H(p)$ and cross-entropy $CE(p, y)$ to the true label, formulated as:
\begin{equation}
    u(p, y) = H(p) + CE(p, y) = -\sum_j{p_j\log p_j} - \log{p_y}
    \label{eq:uncertainty_formula}
\end{equation} 
where $p_j$ is the predicted probability for class $j$, $p_y$ is the predicted probability of the ground truth class $y$, and $m$ denotes the total number of classes. 
This formulation exhibits desirable theoretical properties. First, $u(p, y)$ is non-negative and achieves its minimum value of zero only when the prediction is perfectly correct and confident, i.e., $p_y=1$. Second, this formulation assigns strictly higher uncertainty to incorrect predictions even if they exhibit the same maximum confidence as correct predictions. This property can be analytically demonstrated for both binary and multi-class tasks. 
In the \textbf{binary case $(m=2)$}, when the predicted confidence for the true class is $c$, the uncertainty is given by:
\begin{equation}
    u_{cor}(c) = h(c) - \log{c}
    \label{eq:u_cor_b}
\end{equation}
where $h(c)$ denotes the entropy of the predicted distribution.
If the prediction is incorrect, meaning the predicted confidence for the wrong class is $c$, then the uncertainty becomes:
\begin{equation}
    u_{err}(c) = h(c) - \log{(1-c)}
    \label{eq:u_cor_b}
\end{equation} 
Since $-\log{(1-c)} > -\log{c}$ for $c>0.5$, the incorrect prediction yields a higher uncertainty.
In the \textbf{multi-class case $(m\geq 3)$}, consider a wrong class receiving confidence $c$, and the remaining $(1-c)$ probability being uniformly distributed among $(m-1)$ classes. Then, the minimal uncertainty of such an incorrect prediction is:
\begin{equation}
    u_{err}^{min}(c) = -c\log{c}-(1-c)\log{\frac{1-c}{m-1}}-\log{\frac{1-c}{m-1}}
    \label{eq:u_cor_b}
\end{equation}
In contrast, the uncertainty of a correct prediction with the same confidence $c$ is given by:
\begin{equation}
    u_{cor}(c) = -c\log{c}-(1-c)\log{\frac{1-c}{m-1}}-\log{c}
    \label{eq:u_cor_b}
\end{equation}
The difference between them is expressed as:
\begin{equation}
    u_{err}^{min}(c)-u_{cor}(c) = \log{\frac{c(m-1)}{1-c}}
    \label{eq:u_cor_b}
\end{equation}
This difference is strictly positive when $c>\frac{1}{m}$, theoretically guaranteeing that incorrect predictions always receive greater uncertainty than correct ones with the same maximum confidence. This property enables our uncertainty formulation to effectively penalize confident incorrect predictions.

Based on this uncertainty formulation, for each $k$-view combination subset $C_k^i$, we compute the uncertainty $u_i^k$ of the corresponding prediction $p_k^i$ for $i\!\in\!\{1, 2, \dots , \binom{n}{k}\}$ as follows:
\begin{equation}
    u_k^i = -\sum_{j=1} ^{m} \Bigl[q_j \log(p_k^i(j)) + p_k^i(j) \log(p_k^i(j))\Bigr]
    \label{eq:uncertainty}
\end{equation}
where $p_k^i(j)$ denotes the predicted probability for class $j$ from the $i$-th view combination in $C_k$, and \( q_j\) is the one-hot encoded vector of the ground truth label for class $j$.
To emphasize more reliable predictions, we assign each prediction $p_k^i$ a weight proportional to the inverse of its uncertainty:
\begin{equation}
    w_k^i = \frac{1}{u_k^i}
    \label{eq:weights}.
\end{equation}
In each $k$-view combination set $\mathcal{C}_k$, we then normalize $\{w_k^i\}$ such that $\sum_i w_k^i\!=\!1$.
The combined prediction for each combination set is obtained by taking the weighted sum of individual predictions:
\begin{equation}
    P(\mathcal{C}_k) = \sum_{i} w_k^i \,\cdot\, p_k^i
    \label{eq: prediction for combination}
\end{equation}
This uncertainty-weighted fusion enhances the influence of reliable, low-uncertainty predictions in the mutual distillation process, thereby improving the robustness of training across diverse view combinations. Note that this uncertainty-weighted fusion depends on the ground-truth label $y$ for the cross-entropy term, and therefore it is used only during training. During inference, HMDMV fuses features from all available views and generates a single prediction directly. Since multiple predictions are not produced within a combination set, prediction-level aggregation is not required. The robustness to potentially unreliable views is addressed through the training process itself. By enforcing consistency across all possible view combinations, the model learns more stable predictions even when some views are noisy or occluded.

\subsection{Hierarchical mutual distillation}
\label{sec:HMD}
Previous multi-view mutual distillation methods typically rely on a binary alignment between single-view and full multi-view predictions. This can create a representational gap between local and global contexts and risks overfitting to a potentially noisy fused representation. To mitigate this issue, we introduce Hierarchical Mutual Distillation (HMD), which leverages partial view combinations as intermediate representations. By capturing inter-view correlations, HMD promotes representational robustness against incomplete or variable inputs.
To implement this, we combine multiple loss terms to leverage single-view, partial multi-view, and full multi-view predictions. The total loss function is defined as follows:
\begin{equation}
    L = \sum_{k=1}^{n} L_k 
    \;+\;
    \lambda \,L_{\text{hmd}}
    \label{eq:UWMD_Loss}
\end{equation}
where \( \lambda \) adjusts the contribution of the Hierarchical Mutual Distillation(HMD) loss \( L_{\text{hmd}} \) to the total loss. 
For each $k\in\{1,2,\dots,n\}$, we have a set of $k$-view subsets $\mathcal{C}_k = \{C_k^i\}$ and each subset $C_k^i$ produces a prediction $p_k^i$. We define the mean classification loss across all $k$-view subsets as follows:
\begin{equation}
    L_k = 
    \frac{1}{|\mathcal{C}_k|} 
    \sum_{C_k^i \,\in\, \mathcal{C}_k}
    \mathcal{L}_{\text{CE}}\!\bigl(p_k^i,\,y\bigr)
    \label{eq:L_k_mean}
\end{equation}
where \(\mathcal{L}_{\text{CE}}\) denotes a cross-entropy-based classification loss, and \(y\) is the ground truth label. In particular, $k\!=\!1$ is single-view, $2\!\le\!k\!\le\!n-1$ is partial multi-view, and $k=n$ is full multi-view.

The main component of our method is $L_{\text{hmd}}$, which performs comprehensive mutual knowledge distillation~\cite{zhang2018deep} across all view combination sets. Conventional knowledge distillation~\cite{hinton2015distilling} is a unidirectional transfer process that minimizes the KL-divergence between the temperature-scaled outputs from teacher and student, formulated as:
\begin{equation}
    L_{\text{kd}}(t, s; \tau) = D_{\text{KL}} \left( \tilde{{\sigma}}(t, \tau) \parallel \tilde{{\sigma}}(s, \tau) \right)
    \label{eq: kd_loss}
\end{equation}
where \( t \) and \( s \) are the teacher's and student's logits, and \( \tilde{{\sigma}} \) is the temperature-scaled softmax with hyperparameter \( \tau \) \( (\tau\!>\!0) \).
In \Cref{sec:All_comb} and \Cref{sec:UW}, each $k$-view combination set $\mathcal{C}_k$ produces multiple predictions $\{p_k^i\}$. Through the uncertainty-weighted average of \cref{eq: prediction for combination}, we compute a combined prediction $\hat{p}_k\!=\!\sum_{i} w_k^i \,\cdot\, p_k^i$.
We denote the full multi-view prediction by $p_n$ and define $\hat{p}_{sp}\!\equiv\!\{\hat{p}_1,\,\hat{p}_2,\,\dots,\,\hat{p}_{n-1}\}$ which collectively represents single-view and partial multi-view predictions.
As shown in \Cref{figure_3}, our proposed HMD thus enforces bidirectional knowledge distillation between each $\hat{p}_{sp}$ and $p_n$.
Concretely, the $L_{\text{hmd}}$ is defined as follows:
\begin{equation}
    L_{\text{hmd}} = \sum_{k=1}^{n-1} \frac{1}{2} \tau^2 \left[ L_{\text{kd}}(\hat{p}_{k}, p_n; \tau) + L_{\text{kd}}(p_n, \hat{p}_{k}; \tau) \right]
    \label{eq: hmd_loss}
\end{equation}
where $\tau^2$ is utilized to adjust the gradient magnitude under temperature softening, following Hinton et al.~\cite{hinton2015distilling}. 
As shown in \cref{eq: hmd_loss}, we enhance the previous method~\cite{black2024multi} by introducing sequential mutual distillation between partial and full multi-view predictions, enabling the model to incorporate all combinations of view-wise information more effectively. We treat the full multi-view prediction as the most informative and stable representation and use it as the central reference in the distillation configuration. Each single- and partial-view prediction is, in turn, directly aligned with the full multi-view prediction. As detailed in Supplementary Section C.2, we experimentally compared four configurations: incremental, direct alignment with full multi-view, single-centric, and exhaustive all-pair. The single-centric approach showed the lowest performance, followed by incremental and exhaustive strategies. Based on these results, we adopted direct alignment with the full multi-view prediction, which proved to be the most effective configuration.
We consider the symmetric KL term \(f(p,q)=\tfrac{1}{2}\tau^{2}\!\big[D_{\mathrm{KL}}(p\|q)+D_{\mathrm{KL}}(q\|p)\big]\). In probability space, \(f\) is convex in \((p,q)\) and is uniquely minimized at \(p=q\)~\cite{cover1999elements}. Consequently, the HMD objective \(L_{\mathrm{hmd}}\), defined as a sum of such terms across view levels, drives the partial and full multi-view distributions toward agreement. Unlike unidirectional distillation, our mutual setting updates both distributions simultaneously, and the convexity of $f$ provides a consistent alignment signal for gradient‑based updates. Combined with temperature scaling \(\tau\), which moderates gradient magnitudes in distillation~\cite{hinton2015distilling}, this provides the stable optimization consistently observed across datasets in \Cref{sec:ablation_study}.

In addition, the proposed HMD loss incorporates all partial multi-view combinations, not just single-view. The added loss terms compared to conventional methods can be expressed as:
\begin{equation}
    \Delta=\sum_{k=2}^{n-1} \frac{1}{2} \tau^2 \left[ D_{\text{KL}}(\hat{p}_{k}||p_n) + D_{\text{KL}}(p_n||\hat{p}_{k}) \right]
\end{equation}
where $\Delta$ denotes the additional symmetric KL divergence terms introduced by combining each partial multi-view prediction $\hat{p}_k$ with the full multi-view prediction $p_n$.
This implies that the model learns the relational information between partial and full multi-view predictions that the previous methods fail to capture. As the number of views $n$ increases, these additional terms grow linearly, mitigating information loss. Consequently, the proposed HMD loss not only ensures convexity and training stability but also reduces inter-view prediction discrepancies and enhances robustness under various scenarios.
\begin{algorithm}[t]
\caption{Optimization Algorithm for HMDMV}
\label{alg:hmdmv}

\begin{algorithmic}[1]
    \Require Training data $\mathcal{D}$, epochs $E$, params $\tau, \lambda$
    \Ensure Optimized parameters $\theta$

    \For{epoch $t = 1$ to $E$}
        \For{each mini-batch in $\mathcal{D}$}
            \State Extract features $\mathcal{E}(v)$ for all $n$ views via~\cref{eq:feature_token};
            \For{$k = 1$ to $n$}
                \State \textbf{Forward:} Compute $\{p_k^i, w_k^i\}$ for \textbf{all combinations} $C_k^i \in \mathcal{C}_k$ via \cref{eq:combined_feature_token,eq:prediction,eq:uncertainty,eq:weights};
                \State \textbf{Fusion:} Calculate fused prediction $\hat{p}_k = \sum w_k^i p_k^i$ via \cref{eq: prediction for combination};
            \EndFor
            \State Compute $L_k$ via \cref{eq:L_k_mean} and $L_{hmd}$ between $\{\hat{p}_k\}_{k=1}^{n-1}$ and $p_n$ via \cref{eq: hmd_loss};
            \State Update $\theta$ by minimizing $L = \sum_{k=1}^n{L_{k}} + \lambda L_{hmd}$ via \cref{eq:UWMD_Loss};
        \EndFor
    \EndFor 
\end{algorithmic}
\textbf{Remark:} For scalable training, we propose a \textit{subset sampling} variant in \Cref{sec:scalable_training}.
\end{algorithm}
The overall optimization procedure of HMDMV, which leverages all possible view combinations through feature extraction, uncertainty-weighted fusion, and hierarchical mutual distillation, is summarized in \Cref{alg:hmdmv}.

\section{Experiments}
\label{sec:experiments}
To evaluate our method in both unstructured and structured multi-view scenarios, we conducted experiments on four datasets.
By comparing results, we highlight our method's effectiveness in handling realistic and generalized cases as well as standard multi-view scenarios.
\subsection{Datasets}
\label{sec:dataset}
Hotels-8k~\cite{kamath20212021} consists of 99,513 room images gathered from 7774 different hotels. The room views within the same hotel exhibit minimal to no overlap, effectively representing a highly unstructured multi-view scenario. We use the same training/validation/test split as in \cite{black2024multi}, where hotel IDs are treated as classes for training.
Google Landmarks Dataset V2 (GLDv2)~\cite{weyand2020google} is a large-scale landmark dataset covering both human-made and natural landmarks, captured from diverse angles and conditions. Each landmark is at least partially captured in these images, creating an unstructured multi-view environment. We utilize a subset of 10,000 landmark IDs, comprising 73,956 images for training, 24,354 for validation, and 24,465 for testing.
Carvana~\cite{carvana-image-masking-challenge} dataset contains 6572 vehicles, each with 16 fixed viewpoints. We define 263 vehicle models as classes, splitting 4379, 989, and 989 vehicle IDs into training, validation, and test sets, respectively, with no overlap. 
VinDr-Mammo~\cite{nguyen2023vindr} is a large-scale full-field digital mammography dataset where each exam consists of four standard views: left/right craniocaudal (LCC/RCC) and left/right mediolateral oblique (LMLO/RMLO). From the released cohort, we excluded one outlier exam with abnormal data quality, resulting in 4999 subjects with complete 4-view exams. We formulate a patient-level binary classification task, where an exam is labeled as abnormal if at least one breast is annotated as abnormal, and as normal otherwise. This yields 4082 normal subjects and 917 abnormal subjects. Due to this substantial class imbalance, the dataset was split at the patient-level into training, validation, and test sets while preserving the class distribution in each set. Specifically, we allocated 3599 samples for training, 400 for validation, and 1000 for testing.
For Hotels-8k, GLDv2, and Carvana, images are resized to 224×224, resulting in 49 tokens per view. For VinDr-Mammo, images are resized to 384×384 to preserve fine-grained clinical cues.

\subsection{Experimental setup}
\label{sec:exp_setup}
\begin{figure}[t]
\centering
\includegraphics[width=0.6\columnwidth]{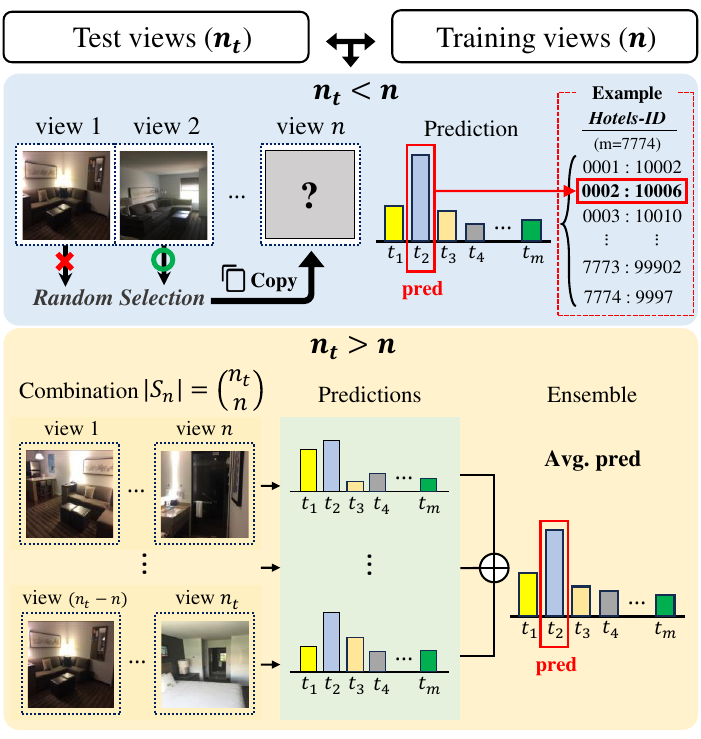}
\caption{Approach for handling view count mismatch between training and test. In realistic scenarios such as unstructured multi-view datasets, the number of test views $n_t$ may differ from the training view count $n$. Since existing multi-view methods cannot directly cope with such conditions, we adopt a unified strategy for fair comparison. When $n_t<n$, the missing views are duplicated to construct $n$-view inputs. When $n_t=n$, inference proceeds identically to training. When $n_t>n$, multiple $n$-view subsets are sampled and their predictions are uniformly averaged. This strategy ensures consistent inference regardless of the number of test views.}
\label{figure_4}
\end{figure}
\noindent \textbf{Implementation details}\hspace{0.2em} Although our method does not require a fixed backbone, we primarily employ a hybrid ResNet26+Small ViT backbone pre-trained on ImageNet~\cite{deng2009imagenet}, with an effective 32$\times$CNN downsampling ratio~\cite{steiner2106train}. For optimization, we use stochastic gradient descent with a warm-restarts (SGDR) scheduler~\cite{loshchilov2016sgdr}, a momentum of 0.9, and a weight decay of $5 \times 10^{-4}$, with a batch size of 64. All experiments are conducted on NVIDIA RTX 6000 Ada and H100 GPUs. For reporting result of test set, we use the checkpoint that achieved the best validation performance. Further details on backbone configurations, such as layer freezing and fine-tuning, are provided in the Supplementary Section A.
\\
\noindent \textbf{Competing methods and parameter settings.}\hspace{0.2em} We compare HMDMV with representative multi-view methods, including MVCNN~\cite{su2015multi}, GVCNN~\cite{feng2018gvcnn}, MVT~\cite{chen2021mvt}, ETMC~\cite{han2022trusted}, and MV-HFMD~\cite{black2024multi}.
For each method, we used the public implementations whenever available; otherwise, we followed
widely-used public reimplementations consistent with the original papers. To ensure a fair comparison, we unified the CNN feature extractor to ResNet-26 for methods where it is configurable. 
MVCNN uses a shared CNN backbone with max pooling across views, optimized with SGD under a constant learning rate.
GVCNN employs a CNN backbone with a view-grouping module and weighted fusion, trained using SGD with a constant learning rate.
MVT adopts a DeiT-S based local--global Transformer with 4 global and 8 local blocks, trained with AdamW using a cosine scheduler.
ETMC integrates a pseudo-view mechanism with an annealing coefficient $\lambda_t$, and is trained with Adam using ReduceLROnPlateau.
MV-HFMD uses an R26+ViT-S backbone and performs mutual distillation between single- and multi-view predictions;
we set the distillation temperature to $\tau=4.0$ and the distillation weight to $\lambda_{\text{HFMD}}=0.1$, and train it using SGD with a one-cycle LR scheduler.
We matched the training budget across all methods, using a batch size of 64 and training for up to 100 epochs using early stopping with a patience of 20. The checkpoint with the best validation performance was selected for evaluation.
Starting from each implementation's recommended settings, we tuned the learning rate via grid search in $[10^{-6}, 10^{-3}]$ on the validation set.
\\
\noindent \textbf{Hyperparameter tuning}\hspace{0.2em} There are three primary hyperparameters for the hierarchical mutual distillation \(\tau\), \(\lambda\), and \(\alpha\), which we optimize to balance the distillation process. We follow the same temperature parameter $\tau=4.0$ as the previous distillation method~\cite{black2024multi}. The distillation weight \(\lambda\) is adaptively adjusted based on the size of each $k$-view combination $(k\!=\!1,\dots,n-1)$ being distilled against the full multi-view. Specifically, we define an initial \(\lambda \), then scale it as follows:
\begin{equation}
    \lambda_k
    \;=\;
    \lambda
    \,\times\,
    k^{\,\alpha},
    \label{eq:lambda_adaptive}
\end{equation}
where $\lambda_k$ is the mutual distillation weight between the $k$-view subset and the full multi-view, and $\alpha$ is an exponent determining how much larger subsets are emphasized.
Based on experiments, we set \(\alpha=1.2\) and \(\lambda=0.1\). Detailed sensitivity analyses for the key hyperparameters are presented in Supplementary Section B. Overall, the method shows stable performance across a reasonable range around the selected values, confirming the robustness of our configuration.
This configuration ensures that as $k$ increases, the $k$-view combination set receives a higher mutual distillation influence, thereby improving the full multi-view predictions.
\\
\noindent \textbf{Evaluation setup} \hspace{0.2em} We adopt evaluation protocols according to whether the dataset is unstructured or structured. For Hotels-8k and GLDv2, we conducted two evaluation settings: (1) using a subset that matches the training view count, and (2) employing a test set where each class may have a different number of views, thus evaluating all possible view combinations for each class. The detailed implementation of the second setting is shown in \Cref{figure_4}. For fairness across methods, we apply the normalized protocol described in \Cref{figure_4} uniformly to all approaches, even though HMDMV supports inference with fewer views without duplication. For the Hotels-8k and GLDv2 datasets, where classes have different numbers of views, we report results separately for both evaluation settings. In contrast, Carvana and VinDr-Mammo follow a structured multi-view protocol with a fixed set of views per sample. Thus, they are evaluated only under the first method using the same view configuration as in training. 
For Hotels-8k, GLDv2, and Carvana, we report Top-1 and Top-5 classification accuracies. For VinDr-Mammo, which is formulated as a patient-level binary classification task with substantial class imbalance, we report AUROC and F1-score. Overall, this unified protocol ensures a fair comparison of multi-view methods under differing test-view conditions across four datasets and highlights our method’s effectiveness in both unstructured and structured multi-view scenarios.
\\
\noindent \textbf{Statistical significance} \hspace{0.2em} We repeat each experiment over multiple random seeds and report mean$\pm$standard deviation. To evaluate the statistical significance of the performance improvements, we conduct two-sided t-tests on the test metrics between HMDMV and each competing method. Statistically significant differences with respect to HMDMV are denoted by $^* (p < 0.05)$, $^{**} (p < 0.01)$, and $^{***} (p < 0.001)$ in the tables, reported only for the primary metrics (Top-1 accuracy, AUROC, and F1-score).
\subsection{Fixed multi-view evaluation}
\label{sec:fixed_multiview_evaluation}
\begin{table}[t]
\centering
\newcommand{\meanstd}[2]{%
  \shortstack{#1{\footnotesize$\pm$#2}}%
}
\resizebox{\columnwidth}{!}{%
\begin{tabular}{lllclclc}
\toprule
\multicolumn{2}{c}{} & \multicolumn{2}{c}{\textbf{2 Views}} & \multicolumn{2}{c}{\textbf{3 Views}} & \multicolumn{2}{c}{\textbf{4 Views}} \\
\cmidrule(lr){3-4} \cmidrule(lr){5-6} \cmidrule(lr){7-8}
\textbf{Dataset} & \textbf{Method} & \multicolumn{1}{c}{\textbf{Top-1}} & \textbf{Top-5} & \multicolumn{1}{c}{\textbf{Top-1}} & \textbf{Top-5} & \multicolumn{1}{c}{\textbf{Top-1}} & \textbf{Top-5} \\
\midrule
\multirow{5}{*}{Hotels-8k~\cite{kamath20212021}}
            & MVCNN~\cite{su2015multi} 
            & \meanstd{41.69}{0.87}$^{***}$ & \meanstd{58.25}{0.81}
            & \meanstd{43.80}{0.62}$^{***}$ & \meanstd{60.86}{0.84}
            & \meanstd{45.61}{0.75}$^{***}$ & \meanstd{62.90}{0.69} \\
            & GVCNN~\cite{feng2018gvcnn}
            & \meanstd{42.01}{0.65}$^{***}$ & \meanstd{57.73}{1.12}
            & \meanstd{44.65}{0.38}$^{***}$ & \meanstd{62.52}{1.09}
            & \meanstd{47.73}{1.01}$^{***}$ & \meanstd{63.18}{0.45} \\
            & MVT~\cite{chen2021mvt} 
            & \meanstd{57.17}{0.32}$^{***}$ & \meanstd{72.54}{0.32}
            & \meanstd{63.92}{0.51}$^{***}$ & \meanstd{78.21}{0.61}
            & \meanstd{67.68}{0.53}$^{***}$ & \meanstd{80.87}{0.75} \\
            & ETMC~\cite{han2022trusted} 
            & \meanstd{51.55}{0.46}$^{***}$ & \meanstd{67.79}{0.49}
            & \meanstd{56.30}{0.37}$^{***}$ & \meanstd{72.09}{0.24}
            & \meanstd{60.79}{0.48}$^{***}$ & \meanstd{76.40}{0.49} \\
            & MV-HFMD~\cite{black2024multi} 
            & \meanstd{64.83}{0.37}$^{*}$ & \meanstd{80.50}{0.42}
            & \meanstd{71.92}{0.41}$^{***}$ & \meanstd{85.24}{0.45} 
            & \meanstd{75.18}{0.60}$^{***}$ & \meanstd{87.99}{0.36} \\
\rowcolor[gray]{.9}
            & \textbf{HMDMV} 
            & \textbf{\meanstd{65.88}{0.45}} & \textbf{\meanstd{80.98}{0.35}}
            & \textbf{\meanstd{74.46}{0.33}} & \textbf{\meanstd{87.46}{0.34}} 
            & \textbf{\meanstd{78.66}{0.42}} & \textbf{\meanstd{90.99}{0.22}} \\
\midrule
\multirow{5}{*}{Google Landmarks Dataset v2~\cite{weyand2020google}}
           & MVCNN~\cite{su2015multi}  
           & \meanstd{85.00}{0.32}$^{***}$ & \meanstd{93.78}{0.33}
           & \meanstd{89.06}{0.64}$^{***}$ & \meanstd{95.37}{0.49}
           & \meanstd{91.14}{0.73}$^{***}$ & \meanstd{96.20}{0.30} \\ 
           & GVCNN~\cite{feng2018gvcnn}
           & \meanstd{86.01}{0.20}$^{***}$ & \meanstd{94.82}{0.36}
           & \meanstd{90.99}{0.22}$^{***}$ & \meanstd{96.73}{0.24}
           & \meanstd{92.91}{0.35}$^{***}$ & \meanstd{97.87}{0.28} \\ 
           & MVT~\cite{chen2021mvt}  
           & \meanstd{91.22}{0.36}$^{***}$ & \meanstd{97.58}{0.22}
           & \meanstd{95.11}{0.28}$^{***}$ & \meanstd{97.56}{0.24}
           & \meanstd{97.17}{0.23}$^{***}$ & \meanstd{99.12}{0.10} \\ 
           & ETMC~\cite{han2022trusted}
           & \meanstd{89.52}{0.45}$^{***}$ & \meanstd{96.88}{0.34}
           & \meanstd{91.87}{0.37}$^{***}$ & \meanstd{97.01}{0.23}
           & \meanstd{93.96}{0.41}$^{***}$ & \meanstd{98.14}{0.20} \\ 
           & MV-HFMD~\cite{black2024multi}
           & \meanstd{96.27}{0.33}$^{**}$ & \meanstd{98.70}{0.17}
           & \meanstd{98.52}{0.14}$^{***}$ & \meanstd{99.69}{0.07}
           & \meanstd{99.83}{0.10} & \meanstd{99.91}{0.07} \\ 
\rowcolor[gray]{.9}
           & \textbf{HMDMV}
           & \textbf{\meanstd{96.82}{0.29}} & \textbf{\meanstd{99.19}{0.13}}
           & \textbf{\meanstd{99.11}{0.08}} & \textbf{\meanstd{99.80}{0.03}}
           & \textbf{\meanstd{99.92}{0.03}} & \textbf{\meanstd{99.97}{0.02}} \\ 
\midrule
\multirow{5}{*}{Carvana~\cite{carvana-image-masking-challenge}}
           & MVCNN~\cite{su2015multi}   
           & \meanstd{90.01}{0.52}$^{***}$ & \meanstd{97.84}{0.33}
           & \meanstd{92.19}{0.78}$^{***}$ & \meanstd{97.94}{0.67}
           & \meanstd{93.50}{0.48}$^{***}$ & \meanstd{98.76}{0.44} \\ 
           & GVCNN~\cite{feng2018gvcnn} 
           & \meanstd{91.46}{0.32}$^{***}$ & \meanstd{98.73}{0.26}
           & \meanstd{93.01}{0.23}$^{***}$ & \meanstd{98.97}{0.36}
           & \meanstd{93.29}{0.48}$^{***}$ & \meanstd{98.83}{0.33} \\
           & MVT~\cite{chen2021mvt}  
           & \meanstd{91.93}{1.20}$^{*}$ & \meanstd{98.81}{0.52}
           & \meanstd{94.22}{0.46}$^{**}$ & \meanstd{98.85}{0.29}
           & \meanstd{94.26}{0.75}$^{**}$ & \meanstd{98.48}{0.25} \\
           & ETMC~\cite{han2022trusted}  
           & \meanstd{90.90}{0.56}$^{**}$ & \meanstd{95.05}{0.45} 
           & \meanstd{91.83}{0.31}$^{***}$ & \meanstd{95.49}{0.25}
           & \meanstd{92.13}{0.33}$^{***}$ & \meanstd{95.41}{0.11} \\
           & MV-HFMD~\cite{black2024multi} 
           & \meanstd{93.39}{0.23} & \meanstd{99.35}{0.09} 
           & \meanstd{94.94}{0.31}$^{**}$ & \meanstd{99.59}{0.10} 
           & \meanstd{95.21}{0.23}$^{*}$ & \meanstd{99.57}{0.08} \\
\rowcolor[gray]{.9}
           & \textbf{HMDMV} 
           & \textbf{\meanstd{93.57}{0.20}} & \textbf{\meanstd{99.49}{0.13}} 
           & \textbf{\meanstd{95.63}{0.36}} & \textbf{\meanstd{99.64}{0.09}} 
           & \textbf{\meanstd{95.87}{0.31}} & \textbf{\meanstd{99.60}{0.00}} \\
\bottomrule
\end{tabular}
}
\caption{Classification performance on multi-view datasets under fixed-view settings. We evaluate existing multi-view methods and our proposed HMDMV on three datasets using the same number of views at test time as during training. For unstructured datasets such as Hotels-8k and GLDv2, we sample exactly $n$ views from identities with at least $n$ images. We report Top-1 and Top-5 accuracies as mean $\pm$ standard deviation over multiple random seeds. Significant differences are denoted by: $^{***} p<0.001$, $^{**} p<0.01$, $^{*} p<0.05$.}
\label{tab:resulttable_1}
\end{table}

\begin{table}[t]
\centering
\small
\newcommand{\meanstd}[2]{%
  \shortstack{#1{\footnotesize$\pm$#2}}%
}
\begin{tabular}{lll}
\toprule
\textbf{Method} & \textbf{AUROC} & \textbf{F1-score} \\ 
\midrule
MVCNN~\cite{su2015multi} & \meanstd{0.734}{0.038}$^{**}$ & \meanstd{0.408}{0.049}$^{**}$ \\
GVCNN~\cite{feng2018gvcnn} & \meanstd{0.735}{0.022}$^{***}$ & \meanstd{0.447}{0.022}$^{**}$ \\
MVT~\cite{chen2021mvt} & \meanstd{0.759}{0.023}$^{**}$ & \meanstd{0.451}{0.011}$^{**}$ \\
ETMC~\cite{han2022trusted} & \meanstd{0.778}{0.022} & \meanstd{0.452}{0.031}$^{*}$ \\
MV-HFMD~\cite{black2024multi} & \meanstd{0.777}{0.023} & \meanstd{0.473}{0.028} \\
\rowcolor[gray]{.9}
\textbf{HMDMV (ours)} & \textbf{\meanstd{0.799}{0.013}} & \textbf{\meanstd{0.497}{0.025}} \\
\bottomrule
\end{tabular}%
\caption{Patient-level binary classification performance on VinDr-Mammo~\cite{nguyen2023vindr} under the standard four-view setting. Due to substantial class imbalance, we report AUROC and F1-score as mean $\pm$ standard deviation. Significant differences are denoted by: $^{**} p<0.01$, $^{*} p<0.05$.}
\label{tab:results_vindr}
\end{table}

We evaluate all methods under the fixed-view protocol, where the view configuration at test time matches the one used during training. For unstructured datasets, each identity contains a variable number of images; thus, we form a fixed-$n$ test instance by sampling exactly $n$ views from identities that provide at least $n$ images. For structured datasets, each sample is associated with a predefined set of views, and we therefore evaluate using the same view configuration as in training. In particular, Carvana~\cite{carvana-image-masking-challenge} provides 16 fixed viewpoints per vehicle, from which we select a subset of $n$ views to reduce redundancy while preserving complementary perspectives. 
We report top-1 and top-5 classification accuracies on Hotels-8k~\cite{kamath20212021}, GLDv2~\cite{weyand2020google}, and Carvana~\cite{carvana-image-masking-challenge} under 2-, 3-, and 4-view settings in \Cref{tab:resulttable_1}.
Overall, our proposed method achieves the strongest performance across most configurations.
On Hotels-8k, HMDMV improves over the recent state-of-the-art approach~\cite{black2024multi} by $+1.05$, $+2.54$, $+3.48\%$ in top-1 accuracy for the 2-, 3-, and 4-view settings, respectively.
On GLDv2, HMDMV is comparable in the 2-view setting but achieves higher accuracy once three or more views are used, suggesting that hierarchical alignment across subset predictions stabilizes fusion when viewpoints are diverse. 
On the Carvana dataset, HMDMV yields consistent gains across view counts, indicating that our approach is beneficial in both unstructured and structured multi-view settings. We further validate HMDMV on VinDr-Mammo~\cite{nguyen2023vindr}, a structured medical image dataset, and summarize the results in \Cref{tab:results_vindr}. HMDMV achieves an AUROC of $0.799\pm0.013$ and an F1-score of $0.497\pm0.025$, outperforming all competing methods. These results suggest that learning from all partial multi-view subsets and enforcing hierarchical consistency leads to more stable and accurate patient-level predictions.

This is expected given the design of HMDMV, as improvements are smaller in the 2-view setting where partial combinations do not exist. Nevertheless, across most fixed-view settings, the improvements of HMDMV over prior methods are statistically significant. Performance improves more noticeably once intermediate subsets ($2 \le k \le n-1$) are available, allowing the model to capture richer inter-view relationships through hierarchical mutual distillation.
\subsection{Unstructured multi-view evaluation}
\label{sec:unstructured_multiview_evaluation}
\begin{table}[t]
\centering
\newcommand{\meanstd}[2]{%
  \shortstack{#1{\footnotesize$\pm$#2}}%
}
\resizebox{\columnwidth}{!}{%
\begin{tabular}{lllclclc}
\toprule
\multicolumn{2}{c}{} & \multicolumn{2}{c}{\textbf{2 Views}} & \multicolumn{2}{c}{\textbf{3 Views}} & \multicolumn{2}{c}{\textbf{4 Views}} \\
\cmidrule(lr){3-4} \cmidrule(lr){5-6} \cmidrule(lr){7-8}
\textbf{Dataset} & \textbf{Method} & \multicolumn{1}{c}{\textbf{Top-1}} & \textbf{Top-5} & \multicolumn{1}{c}{\textbf{Top-1}} & \textbf{Top-5} & \multicolumn{1}{c}{\textbf{Top-1}} & \textbf{Top-5} \\
\midrule
\multirow{5}{*}{Hotels-8k~\cite{kamath20212021}}
            & MVCNN~\cite{su2015multi}
            & \meanstd{49.06}{0.45}$^{***}$ & \meanstd{63.74}{0.82}
            & \meanstd{43.87}{0.82}$^{***}$ & \meanstd{59.86}{0.56}
            & \meanstd{39.09}{0.46}$^{***}$ & \meanstd{55.06}{0.63} \\
            & GVCNN~\cite{feng2018gvcnn}
            & \meanstd{50.18}{1.11}$^{***}$ & \meanstd{64.90}{0.90}
            & \meanstd{45.06}{0.77}$^{***}$ & \meanstd{60.44}{0.42}
            & \meanstd{39.93}{0.56}$^{***}$ & \meanstd{55.49}{0.88} \\
            & MVT~\cite{chen2021mvt}
            & \meanstd{64.89}{0.51}$^{***}$ & \meanstd{78.27}{0.35}
            & \meanstd{61.59}{0.22}$^{***}$ & \meanstd{75.41}{0.61}
            & \meanstd{57.49}{0.43}$^{***}$ & \meanstd{71.37}{0.41} \\
            & ETMC~\cite{han2022trusted}
            & \meanstd{59.28}{0.45}$^{***}$ & \meanstd{72.29}{0.44}
            & \meanstd{55.64}{0.52}$^{***}$ & \meanstd{69.96}{0.76}
            & \meanstd{52.19}{0.64}$^{***}$ & \meanstd{66.77}{0.51} \\
            & MV-HFMD~\cite{black2024multi} 
            & \meanstd{70.96}{0.25}$^{***}$ & \meanstd{84.82}{0.33} 
            & \meanstd{71.18}{0.39}$^{**}$ & \meanstd{84.21}{0.39} 
            & \meanstd{69.67}{0.64}$^{**}$ & \meanstd{83.45}{0.33} \\
\rowcolor[gray]{.9}
            & \textbf{HMDMV} 
            & \textbf{\meanstd{72.02}{0.27}} & \textbf{\meanstd{85.22}{0.23}} 
            & \textbf{\meanstd{72.96}{0.38}} & \textbf{\meanstd{86.12}{0.45}} 
            & \textbf{\meanstd{72.66}{0.51}} & \textbf{\meanstd{85.73}{0.64}} \\
\midrule
\multirow{5}{*}{Google Landmarks Dataset v2~\cite{weyand2020google}}
           & MVCNN~\cite{su2015multi}   
           & \meanstd{83.47}{0.48}$^{***}$ & \meanstd{91.72}{0.68}
           & \meanstd{77.39}{0.71}$^{***}$ & \meanstd{87.43}{0.62}
           & \meanstd{67.66}{0.41}$^{***}$ & \meanstd{78.48}{0.47} \\ 
           & GVCNN~\cite{feng2018gvcnn} 
           & \meanstd{84.89}{0.53}$^{***}$ & \meanstd{92.73}{0.52}
           & \meanstd{79.76}{0.78}$^{***}$ & \meanstd{89.55}{0.52}
           & \meanstd{71.34}{0.76}$^{***}$ & \meanstd{82.86}{0.47} \\ 
           & MVT~\cite{chen2021mvt}  
           & \meanstd{91.99}{0.56}$^{*}$ & \meanstd{95.85}{0.68}
           & \meanstd{89.84}{0.55}$^{***}$ & \meanstd{93.71}{0.38}
           & \meanstd{85.33}{0.74}$^{***}$ & \meanstd{89.47}{0.84} \\ 
           & ETMC~\cite{han2022trusted}
           & \meanstd{89.42}{0.43}$^{***}$ & \meanstd{93.50}{0.75}
           & \meanstd{84.80}{0.57}$^{***}$ & \meanstd{89.51}{0.59}
           & \meanstd{79.02}{0.59}$^{***}$ & \meanstd{85.44}{0.57} \\ 
           & MV-HFMD~\cite{black2024multi} 
           & \meanstd{93.73}{0.60} & \meanstd{97.68}{0.28}
           & \meanstd{94.11}{0.22}$^{**}$ & \meanstd{97.63}{0.21}
           & \meanstd{93.52}{0.49}$^{**}$ & \meanstd{97.49}{0.20} \\ 
\rowcolor[gray]{.9}
           & \textbf{HMDMV} 
           & \textbf{\meanstd{94.09}{0.74}} & \textbf{\meanstd{97.80}{0.21}} 
           & \textbf{\meanstd{94.91}{0.15}} & \textbf{\meanstd{98.11}{0.12}} 
           & \textbf{\meanstd{94.64}{0.12}} & \textbf{\meanstd{97.97}{0.14}} \\
\bottomrule
\end{tabular}
}
\caption{Classification performance on unstructured multi-view dataset. This evaluation setting reflects more realistic scenarios where each test sample can contain a varying number of views. We compare existing multi-view methods and our proposed HMDMV trained with 2, 3, or 4 views. We report as mean $\pm$ standard deviation, and significance marks follow; $^{***} p<0.001$; $^{**} p<0.01$; $^{*} p<0.05$.}
\label{tab:resulttable_2}
\end{table}
In this evaluation, we adapt to the actual number of views available for each class, thereby fully reflecting the intrinsic variability of unstructured datasets. As shown in \Cref{figure_4}, if a class has fewer view images than the training-time count during inference, certain images are randomly duplicated for interpolation. If a class has more view images, we enumerate all valid combinations based on the training-time view count, ensemble the resulting predictions, and compute the final performance. This protocol is applied consistently to all compared methods that are otherwise unable to properly evaluate unstructured multi-view scenarios. This approach allows every available image to contribute to the classification, thereby simulating realistic scenarios in which classes (e.g., hotel rooms or landmarks) may have widely varying numbers of views.

\Cref{tab:resulttable_2} presents the top-1 and top-5 accuracies for the unstructured multi-view datasets Hotels-8k~\cite{kamath20212021} and GLDv2~\cite{weyand2020google} under the 2-, 3-, and 4-view settings. 
Across both datasets, HMDMV consistently surpasses existing multi-view methods. On Hotels-8k, it outperforms the latest method~\cite{black2024multi} by +1.06\%, +1.78\%, and +2.99\% in top-1 accuracy as the number of views increases, with the margin widening at higher view counts. On GLDv2, HMDMV also demonstrates superior performance across all view settings. Unlike the fixed multi-view evaluation, where accuracy generally improves with additional views, the unstructured setting often leads to unstable or even degraded performance for prior methods as the number of views increases. In contrast, HMDMV remains stable, either improving or maintaining accuracy as views increase. These results highlight the robustness of HMDMV in unstructured scenarios where each class or object may contain fewer or more views than those used during training.
\subsection{Ablation study}
\label{sec:ablation_study}
\begin{table}[t]
\centering
\begin{tabular}{cccccc}
\toprule
\makecell{\textbf{Partial} \\ \textbf{Multi-view}} & \makecell{\textbf{Uncertainty-based} \\ \textbf{Weighting}} & \makecell{\textbf{HMD Loss} \\ \textbf{$L_{\text{hmd}}$}} & \makecell{\textbf{Adaptive} \\ \textbf{$\lambda$}} & \textbf{Top-1} & \textbf{Top-5} \\
\midrule
 &  &  &  & 70.39\% & 83.72\% \\
\checkmark &  &  &  & 71.25\% & 84.29\% \\
\checkmark & & \checkmark & & 72.28\% & 85.85\% \\
\checkmark & \checkmark & \checkmark & & 72.52\% & 86.01\% \\
\checkmark & & \checkmark & \checkmark & 72.38\% & 85.84\% \\
\rowcolor[gray]{.9}
\checkmark & \checkmark & \checkmark & \checkmark & \textbf{72.96\%} & \textbf{86.12\%} \\
\bottomrule
\end{tabular}
\caption{%
Ablation study of HMDMV components. We enable each component (\(\checkmark\)) such as partial multi-view, uncertainty weighting, adaptive \(\lambda\), and $L_{\text{hmd}}$. Results are reported on the Hotels-8k dataset under the 3-view training setting, showing Top-1 and Top-5 classification accuracies for each configuration and their combined effect.}
\label{tab:ablation_study}
\end{table}
We evaluate the contributions of the main components of the proposed HMDMV: partial multi-view, uncertainty weighting, hierarchical mutual distillation (HMD) loss $L_{hmd}$, and the adaptive hyperparameter $\lambda$. As the baseline, we adopt a method that excludes all additional components. \Cref{tab:ablation_study} summarizes the results on the Hotels-8k dataset under the 3-view training setting. 

Introducing partial multi-view training, which updates the losses of view combinations, enables more accurate capture of inter-view relationships, yielding a 0.86\% improvement in top-1 accuracy. Adding HMD loss $L_{hmd}$ for enforcing prediction consistency across view combinations further improves performance by 1.13\%. Within the HMDMV framework, applying uncertainty-based weighting and the adaptive $\lambda$ into the HMD-only configuration provides additional improvements of 0.24\% and 0.10\%, respectively. Finally, the complete method integrating all components achieves 2.57\% higher top-1 accuracy and 2.40\% higher top-5 accuracy compared to the baseline. These results demonstrate that each component contributes in a complementary manner, thereby enhancing the robustness of HMDMV in realistic multi-view scenarios. In Supplementary Section C.1, we further analyze the effect of different backbone architectures under the HMD loss $L_{hmd}$, which consistently shows performance improvements across model scales.
\subsection{Scalable training via view-combination subset sampling}
\label{sec:scalable_training}
\begin{table}[t]
\centering
\resizebox{\columnwidth}{!}{%
\begin{tabular}{ccccc@{\hspace{6mm}}ccc}
\toprule
 \textbf{Training view} & \textbf{Subset $M$} & \textbf{Method} & \textbf{Top-1} {$\bm{\uparrow}$} & \textbf{Top-5} {$\bm{\uparrow}$} & \textbf{Train time (s) {$\bm{\downarrow}$}} & \textbf{Peak Memory (GB) {$\bm{\downarrow}$}} & \textbf{Relative cost(\%) {$\bm{\downarrow}$}} \\
\midrule
\multirow{5}{*}{\textbf{3 Views}} 
& \multirow{2}{*}{$M=1$} 
& RS w/o repl & 71.42\% & 84.53\% & \multirow{2}{*}{\textbf{361.55}} & \multirow{2}{*}{\textbf{21.10}} & \multirow{2}{*}{\textbf{74.86}} \\
& 
& RS w/ repl  & 71.65\% & 84.70\% \\
\cmidrule(lr){2-8} 
& \multirow{2}{*}{$M=2$} 
& RS w/o repl & 72.34\% & 85.15\% & \multirow{2}{*}{421.17} & \multirow{2}{*}{23.11} & \multirow{2}{*}{87.21}\\
& 
& RS w/ repl  & \underline{72.52\%} & \underline{85.89\%} \\
\cmidrule(lr){2-8} 
& Exhaustive & None & \textbf{72.96}\% & \textbf{86.12}\% & 482.96 & 25.12 & 100 \\
\specialrule{0.8pt}{1pt}{1pt}
\multirow{9}{*}{\textbf{4 Views}} 
& \multirow{2}{*}{$M=1$} 
& RS w/o repl & 71.57\% & 84.62\% & \multirow{2}{*}{\textbf{521.03}} & \multirow{2}{*}{\textbf{30.54}} & \multirow{2}{*}{\textbf{48.33}} \\
& 
& RS w/ repl  & 71.86\% & 85.12\% \\
\cmidrule(lr){2-8} 
& \multirow{2}{*}{$M=2$} 
& RS w/o repl & 72.06\% & 85.25\% & \multirow{2}{*}{667.26} & \multirow{2}{*}{35.62} & \multirow{2}{*}{61.89} \\
& 
& RS w/ repl  & 72.34\% & \underline{85.51\%} \\
\cmidrule(lr){2-8} 
& \multirow{2}{*}{$M=3$} 
& RS w/o repl & 72.10\% & 85.34\% & \multirow{2}{*}{808.61} & \multirow{2}{*}{40.72} & \multirow{2}{*}{75.00} \\
& 
& RS w/ repl  & \underline{72.38\%} & 85.46\% \\
\cmidrule(lr){2-8} 
& \multirow{2}{*}{$M=4$} 
& RS w/o repl & 72.12\% & 85.40\% & \multirow{2}{*}{951.54} & \multirow{2}{*}{45.82} & \multirow{2}{*}{88.26} \\    
& 
& RS w/ repl  & 72.15\% & 85.48\% \\
\cmidrule(lr){2-8} 
& Exhaustive & None & \textbf{72.66}\% & \textbf{85.73}\% & 1078.13 & 49.82 & 100 \\
\bottomrule
\end{tabular}
}
\caption{Evaluation of pre-fusion view-combination subset sampling strategies and training cost. We compare random sampling without replacement (RS w/o repl) and with replacement (RS w/ repl) against exhaustive training using all view combinations each epoch on Hotels-8k under 3- and 4-view settings. At each epoch, $M$ denotes the number of partial-view subsets resampled per $k$-view configuration. Reported metrics are Top-1/Top-5 accuracy, per-epoch training time, peak memory, and relative computational cost. Relative computational cost is normalized so that exhaustive training equals 100.}
\label{tab:rs_ablation}
\end{table}

The number of possible view combinations increases exponentially with the number of input views $n$, imposing significant computational overhead on HMDMV. In particular, training with all $2^n-1$ view combinations incurs a complexity of $\mathcal{O}(2^n)$, which makes it impractical for multi-view tasks involving a large number of views.
To address this scalability issue, we propose a repeated random subset sampling strategy~\cite{okanovic2023repeated} that reduces computational cost while preserving the diversity of view combinations essential for effective learning. Specifically, single-view and full multi-view combinations are always included during training, while the number of partial multi-view combinations is restricted through sampling. For each partial multi-view level $k (2 \leq k \leq n-1)$, we randomly sample $m_k$ subsets defined as:
\begin{equation}
    m_k = \min\left( M, \binom{n}{k} \right)
    \label{eq: random_m}
\end{equation}
where $M$ is a hyperparameter that controls the maximum number of sampled subsets at each level. This enables flexible control over the number of combinations used for training, independent of $n$. 
Under this strategy, the number of combinations per epoch is given by:
\begin{equation}
    N_{\text{iter}} = n + 1 + \sum_{k=2}^{n-1} m_k \approx n + M(n-2) 
    \Rightarrow T_{\text{sample}} = \mathcal{O}(Mn)
\label{eq:subset_complexity}
\end{equation}
This reduces the original exponential complexity to a linear order, thereby enabling efficient and scalable training across varying numbers of views.

We evaluate two variants of this strategy within HMDMV on the Hotels-8k dataset under the 3- and 4-view training settings. Random sampling without replacement minimizes the likelihood of selecting samples that were chosen in the previous epoch again. In contrast, random sampling with replacement reinitializes the sampling probability at each epoch, ensuring independence between epochs.
As shown in \Cref{tab:rs_ablation}, the repeated random sampling with replacement strategy consistently outperformed its without-replacement strategy counterpart across all performance metrics. Excluding exhaustive training over all possible combinations, the best top-1 accuracies in the 3- and 4-view training settings were achieved with $M=2$ and $M=3$, respectively. Although these results are 0.44\% and 0.28\% lower than those achieved with exhaustive training over all possible combinations, they still surpass the previous state-of-the-art method~\cite{black2024multi} by a substantial margin. Moreover, the proposed sampling strategy is significantly more efficient in terms of training time and peak memory usage compared to the exhaustive variant, as further detailed in \Cref{sec:Computational_cost}.
These results demonstrate that substantial performance improvements can be achieved through random subset sampling at each epoch, even without training on all possible combinations. This strongly suggests the scalability and potential of the proposed HMDMV.
\subsection{Computational cost analysis}
\label{sec:Computational_cost}
\begin{table}[t]
\centering
\small
\begin{tabular}{l c c c c}
\toprule
Method & \makecell{Params {$\bm{\downarrow}$} \\ (M)} & \makecell{Latency {$\bm{\downarrow}$} \\ (ms)} & \makecell{Throughput {$\bm{\uparrow}$} \\ (samples/s)} & \makecell{FLOPs {$\bm{\downarrow}$} \\ (G)} \\
\midrule
MVCNN       & 15.99    & 3.28 & 304.01$\pm$0.17 & 14.06 \\
GVCNN       & 29.88    & 4.32 & 231.63$\pm$0.54 & 14.09  \\
MVT         & 24.65    & 2.10 & 481.69$\pm$1.08 & 27.61  \\
ETMC        & 90.25    & 5.07 & 197.39$\pm$3.26 & 14.19  \\
MV\mbox{-}HFMD & 39.04 & 8.21 & 121.75$\pm$0.39 & 29.28  \\
\rowcolor{gray!10}\textbf{HMDMV} &
39.04 & 6.44 & 152.94$\pm$0.82 & 14.74 \\
\bottomrule
\end{tabular}
\caption{Computational cost of multi-view methods in inference. Params (M): total learnable parameters (millions). Latency (ms): per-sample latency. Throughput (samples/s): throughput, reported as mean $\pm$ SD over $N$ runs. FLOPs (G): operations per forward pass with 3 views. Results are obtained in the 3-view setting, including HMDMV. Note that in inference, HMDMV does not depend on how many view combinations were used during training.}
\label{tab:inf_efficiency_3view}
\end{table}
We analyze the computational cost of HMDMV in terms of training and inference. Let $n$ denote the number of input views, $S$ the number of spatial tokens per view, and $d$ the embedding dimension.

\textbf{Training complexity.} During the exhaustive training phase, HMDMV processes all $(2^n-1)$ view combinations. For each $k$-view combination, the Transformer's self-attention mechanism operates on $kS$ tokens with complexity $\mathcal{O}((kS)^2d)$. By summing over all possible combinations, the total training complexity becomes $\mathcal{O}(2^n \cdot n^2S^2d)$. With the random subset sampling (\Cref{sec:scalable_training}), we restrict the number of subsets to a constant $M$ per level. Consequently, the complexity is significantly reduced from exponential $\mathcal{O}(2^n)$ to linear $\mathcal{O}(Mn)$ with respect to the view count, ensuring scalability. We empirically validate this analysis in \Cref{tab:rs_ablation}. In the 3-view setting, the best-performing configuration is obtained with $M=2$, which reduces training time by 12.79\% and peak memory usage by 2.01GB. In the 4-view setting, the best-performing is achieved with $M=3$, reducing training time by 25\% and peak memory usage by 9.2GB. These results indicate that modest accuracy losses are offset by substantial efficiency gains, supporting the scalability of HMDMV to multi-view tasks with a large number of input views.

\textbf{Inference complexity.} Unlike training, the inference phase is highly efficient as it does not require enumerating view combinations. HMDMV simply extracts features from the given $n$ views and performs a single pass using the fused full multi-view feature. Thus, the inference complexity is $\mathcal{O}(n \cdot C_{backbone}+(nS)^2d)$ where $C_{backbone}$ is the cost of the shared CNN encoder. As shown in \Cref{tab:inf_efficiency_3view} under the 3-view setting, HMDMV maintains the same parameters (39.04M) as the state-of-the-art method~\cite{black2024multi}, while reducing FLOPs from 29.28G to 14.74G, a decrease of approximately 49.6\%. Latency is also reduced from 8.21ms to 6.44ms, corresponding to an improvement of about 21.6\%. Furthermore, throughput improves by around 25.6\%. These results clearly demonstrate that HMDMV achieves enhanced inference efficiency without incurring the computational overhead associated with multi-view training.
\subsection{Inference flexibility}
\label{sec:Flexibility}
\begin{table}[t]
\centering
\small
\begin{tabular}{l c c c}
\toprule
& & \multicolumn{2}{c}{\textbf{Flexibility in inference}} \\
\cmidrule(lr){3-4}
\textbf{Method} &\makecell{\textbf{Flexible view angle} \\ \textbf{across samples}} & \makecell{\textbf{More views} \\ \textbf{than used in training}} & \makecell{\textbf{Fewer views} \\ \textbf{than used in training}} \\
\midrule 
        MVCNN
        & \ding{55}
        & \ding{55} 
        & \ding{55} \\
        GVCNN 
        & \ding{55} 
        & \ding{55}
        & \ding{55} \\
        MVT 
        & \ding{55}
        & \ding{55}
        & \ding{55} \\
        ETMC
        & \ding{55}
        & \ding{55}
        & \ding{55} \\
        MV-HFMD  
        & \checkmark
        & \ding{55}
        & \(\triangle\) \\
        \rowcolor[gray]{.9}
        HMDMV
        & \checkmark 
        & \checkmark 
        & \checkmark \\
\bottomrule
\end{tabular}
\caption{Inference flexibility of multi-view fusion methods. HMDMV has practical advantages in terms of flexible view angle across samples (in training and test) and flexible view count in inference ($n_t\!\le\!n$ or $n_t\!>\!n$).
Symbols denote: \checkmark (fully supported), \(\triangle\) (partially supported), and \ding{55} (not supported).
}
\label{tab:method_comparsion}
\end{table}
\begin{table*}[t]
\centering
\small
  %
  \begin{tabular}{ccccccccc}
    \toprule
    & \multicolumn{8}{c}{\textbf{Inference Views $(n_t)$}} \\
    \cmidrule(lr){2-9}
    \multirow{2}{*}{\textbf{Training Views $(n)$}}
      & \multicolumn{2}{c}{\textbf{1}} & \multicolumn{2}{c}{\textbf{2}} & \multicolumn{2}{c}{\textbf{3}} & \multicolumn{2}{c}{\textbf{4}} \\
     \cmidrule(lr){2-3} \cmidrule(lr){4-5} \cmidrule(lr){6-7} \cmidrule(lr){8-9}
     & Top-1 & Top-5 & Top-1 & Top-5 & Top-1 & Top-5 & Top-1 & Top-5 \\
    \midrule
    1 & 44.56\% & 62.39\% & 56.09\% & 74.11\% & 61.71\% & 79.13\% & 66.13\% & 83.18\% \\
    2 & 48.16\% & 65.16\% & 65.88\% & 80.98\% & 72.58\% & 86.52\% & 76.84\% & 90.13\% \\
    3 & 51.40\% & 68.68\% & 67.15\% & 82.76\% & 74.46\% & 87.46\% & 77.92\% & 90.59\% \\
    4 & \textbf{52.88\%} & \textbf{69.89\%} & \textbf{67.94\%} & \textbf{83.14\%} & \textbf{74.97\%} & \textbf{88.15\%} & \textbf{78.66\%} & \textbf{90.99\%} \\
    \bottomrule
  \end{tabular}
  \caption{Performance of HMDMV with varying training and inference view counts. Each row indicates the number of views used for training, while each column shows the number of views used for inference. We conducted the experiment on the Hotels-8k~\cite{kamath20212021} dataset's fixed subset.}
  \label{tab:flexible_inference}
\end{table*}

In more realistic scenarios, it is often difficult to secure the same number of views for each object, and the available views may vary due to perspective, illumination, or occlusion. Thus, robust multi-view classification requires a method that can flexibly handle such variations. As shown in \Cref{tab:method_comparsion}, previous methods such as MVCNN~\cite{su2015multi}, GVCNN~\cite{feng2018gvcnn}, MVT~\cite{chen2021mvt}, and ETMC~\cite{han2022trusted} assume a fixed number of views, making them unable to handle mismatches between training and inference. Although MV-HFMD~\cite{black2024multi} considers unstructured cases, it supports inference only with a single view or exactly the training view count, which limits flexibility. In contrast, HMDMV can directly process unstructured multi-view inputs with varying numbers and angles of views without additional preprocessing.
Specifically, when fewer views are available at inference than during training, HMDMV processes them directly without replication or padding. This is possible because training already incorporated all subsets from one to $k$ views, where $k$ denotes the maximum number of training views. Conversely, when more views are provided, HMDMV performs inference by ensembling predictions across all valid subsets of size $k$.

The results in \Cref{tab:flexible_inference} further support this flexibility, revealing two consistent trends. (i) When the number of training views $n$ is fixed, accuracy steadily improves as the number of inference views $n_t$ increases, reflecting the benefits of additional information. (ii) When $n_t$ is fixed, models trained with more views achieve higher accuracy, as more informative inter-view representations are trained. Therefore, in practical applications, it is effective for HMDMV to be trained with as many views as possible and to be provided with as many views as available for each object at inference.


\section{Conclusion}
\label{sec:conclusion}
In this study, we proposed the HMDMV framework, which is effective in both structured and unstructured multi-view environments. The method leverages all possible view combinations during training to capture inter-view relationships and incorporates uncertainty-based weighting with hierarchical mutual distillation to achieve robust representation learning. Experiments on diverse large-scale multi-view datasets demonstrated that HMDMV consistently outperforms state-of-the-art methods. Moreover, the framework exhibits flexibility at inference when the number of available views varies. We further observe that HMDMV typically yields larger relative improvements in unstructured multi-view scenarios than in structured fixed-view settings. In structured datasets, views are captured under controlled and consistent viewpoints, so discriminative cues are more easily aligned across views and existing methods already approach a performance ceiling. In contrast, unstructured data exhibit substantial heterogeneity in view count, coverage, and quality across samples. In such cases, adding more views may introduce redundancy, noise, or misleading observations. By treating partial multi-view subsets as intermediate contexts and enforcing hierarchical consistency while reducing the influence of unreliable combinations during training, HMDMV can better exploit additional views and remain robust as the view count increases.

HMDMV has an inherent limitation in that considering all possible view combinations during training leads to an exponential increase in computational cost as the number of views grows. To alleviate this issue, we introduced a repeated random subset sampling strategy, which substantially reduces training time and memory usage while incurring only minor accuracy losses. These results confirm that HMDMV is scalable and practical for large-scale multi-view learning scenarios.
Future work will focus on improving scalability and extending applicability. First, while repeated random subset sampling substantially reduces the combinatorial training cost, it treats partial multi-view combinations uniformly and does not prioritize the most informative combinations. We will investigate adaptive subset selection strategies that leverage uncertainty- and diversity-aware criteria to emphasize informative combinations. This approach aims to prune redundant or noisy subsets and further reduce the training cost for a target accuracy. Second, we will extend HMDMV beyond supervised multi-view classification to broader settings, including retrieval and metric learning, detection and segmentation, and semi- or self-supervised learning under incomplete views. We also plan to adapt the framework to multi-modal learning by systematically integrating heterogeneous modalities such as visual, textual, and auditory data. In this direction, the hierarchical distillation principle will be generalized to multi-modal fusion by treating each modality as a view and distilling across modality subsets, enabling robustness under missing-modality conditions.

In conclusion, the proposed HMDMV framework demonstrates the capability to achieve accuracy, flexibility, and robustness in multi-view learning. We believe that the proposed framework provides not only a practical solution for real-world multi-view applications but also an innovative perspective that can guide future research on multi-view learning and multi-modal fusion.
\section*{Author contributions: CRediT}
\textbf{Yang, J.: }Formal Analysis, Investigation, Data curation, Software, Validation, Visualization, Methodology, Writing – original draft, Writing – Review \& Editing.  \textbf{Chung, H.: }Funding acquisition, Resources, Writing – Review \& Editing. \textbf{Jang, I.: } Conceptualization, Methodology, Formal Analysis, Funding acquisition, Resources, Supervision, Writing – Review \& Editing.

\section*{Acknowledgements}
The authors appreciate Junho Moon and Daehwan Kim for providing valuable feedback during manuscript preparation. 
This work was supported by the National Institute of Health (NIH) research projects (2024ER040700, 2025ER040300), the National Research Foundation of Korea (NRF) grants funded by the Ministry of Science and ICT (MSIT) (RS-2024-00455720, RS-2024-00338048, RS-2024-00414119), the National Supercomputing Center with supercomputing resources including technical support (KSC-2024-CRE-0021, KSC-2025-CRE-0065), the High-Performance Computing Support project (RQT-25-070083) funded by MSIT, a grant of the Korea Health Technology R\&D Project through the Korea Health Industry Development Institute (KHIDI), funded by the Ministry of Health \& Welfare (RS-2025-02220534), the Technology Innovation Program (or Industrial Strategic Technology Development Program - Biotechnology)(RS-2025-13002970) funded By the Ministry of Trade Industry \& Energy (MOTIE, Korea), the Technology Development Program (RS-2024-00510461) funded by the Ministry of SMEs and Startups(MSS, Korea), and Hankuk University of Foreign Studies Research Fund of 2025. 
The work was also supported by Culture, Sports and Tourism R\&D Program through the Korea Creative Content Agency grant funded by the Ministry of Culture, Sports and Tourism (RS-2024-00332210), Artificial Intelligence Graduate School Program (RS-2020-II201373, Hanyang University) supervised by the IITP, and under the artificial intelligence semiconductor support program to nurture the best talents ((IITP-(2025)-RS-2023-00253914) grant funded by the Korea government.

\section*{Declaration of generative AI and AI-assisted technologies in the writing process}
During the preparation of this work, the author(s) used ChatGPT only to improve grammatical precision and refine sentence-level clarity. After using this tool/service, the author(s) reviewed and edited the content as needed and take(s) full responsibility for the content of the publication.

\bibliographystyle{elsarticle-num} 
\bibliography{cas-refs_revision}

\end{document}